%% file: main.tex
\documentclass{article}
\usepackage{booktabs}
\usepackage{graphicx}
\usepackage{amsfonts}
\usepackage{amsmath}
\usepackage{bm}
\usepackage{float}
\usepackage{wrapfig}
\usepackage{hyperref}
\usepackage[preprint]{corl_2026}

\title{Steering Autoregressive Vision-Language-Action Policies via Action Token Intervention}

\author{
  Jason Chan \\
  Department of Computer Science \\
  University of California, Los Angeles \\
  \texttt{jasontchan@ucla.edu} \\
  \And
  Jonathan C. Kao \\
  Departments of Electrical and Computer Engineering, \\
  Computer Science, and Neurobiology\\
  University of California, Los Angeles \\
  \texttt{kao@seas.ucla.edu} \\
}

\begin{document}
\maketitle

\begin{abstract}

We present Token Steering (TS), a method for dynamically steering trajectories generated by an autoregressive vision-language-action (VLA) model through direct intervention in the action-token space.
TS injects low-dimensional user inputs into the model's native action-token representation, allowing users to influence trajectory generation without modifying the underlying vision-language model (VLM) architecture.
Because TS operates entirely at inference time, it requires no additional training or finetuning. 
User inputs guide rather than override the pretrained policy, allowing users to influence robot actions while preserving the dexterity, smoothness, and task priors learned by the VLA. 
We evaluate TS on two household manipulation tasks---drawer closing after object placement and state-aware object swapping---and improve success rates from 10.0\% to 72.5\% and from 16.7\% to 93.8\%, respectively. 
By enabling lightweight, intuitive steering over robot foundation models, our interface has the potential to improve human-robot interaction in consumer environments and broaden accessibility for individuals with limited physical control. Project website: \href{https://jasontchan.github.io/token-steering/}{this https URL}.
    
\end{abstract}


\section{Introduction}

Vision-language-action (VLA) policies demonstrate strong generalization capability by combining large-scale vision-language pretraining together with robot trajectories~\citep{brohan_rt-1_2023, brohan_rt-2_2023, driess_palm-e_2023, kim_openvla_2024, black_pi_0_2024, octo_2023}. 
Recent VLAs, including RT-2, OpenVLA, and $\pi_0$ learn to synthesize robot behavior conditioned on multimodal observations, enabling robots to perform a wide range of manipulation tasks in open-world environments~\citep{brohan_rt-2_2023, kim_openvla_2024, black_pi_0_2024}. 
Despite rapid progress, grounding language in the physical world~\citep{huang_inner_2022, huang_voxposer_2023, driess_palm-e_2023, gao_2026_taxonomy}, long-horizon reasoning and execution~\citep{driess_palm-e_2023, huang_inner_2022, intelligence_pi_05_2025}, and real-time adaptation during deployment remain active areas of research for modern VLA systems.
In practice, failures often arise from small trajectory-level errors that humans could immediately recognize and correct.

Existing interfaces for autonomous guidance rely primarily on language prompting~\citep{ahn_do_2022, liang_code_2022, brohan_rt-2_2023, kim_openvla_2024, black_pi_0_2024, intelligence_pi_05_2025} or full teleoperation~\citep{khazatsky_droid_2025}. 
However, language is poorly suited for fine-grained real-time trajectory correction, while direct teleoperation removes the autonomous capability learned by the policy. 
This creates a fundamental challenge for steerable systems: how can users influence robot behavior during rollouts while preserving the policy’s learned dexterity and task priors?

\input{fig1}

We propose Token Steering (TS), an inference-time method for steering autoregressive vision-language-action policies through direct action-token modification. 
TS exploits the discrete autoregressive structure exposed by tokenized action generation.
Rather than replacing robot trajectories through low-level control, TS injects user-generated steering tokens into the autoregressive action sequence.
The policy then completes the remaining trajectory conditioned on the modified token prefix, preserving smoothness, task consistency, and environmental adaptation. 
We demonstrate TS using $\pi_0$-FAST~\citep{pertsch_fast_2025}, an autoregressive VLA that represents robot trajectories using frequency-domain (FAST) action tokens. 

Overall, we  demonstrate three key findings. 
First, partial token injection enables controllable steering while preserving autonomous VLA trajectory generation. 
Second, editing low-frequency FAST tokens disproportionately influences global trajectory semantics, while higher-frequency tokens primarily affect local refinements. 
Third, TS enables users to correct grounding failures and complete long-horizon state-dependent tasks where a baseline autoregressive VLA fails.
Together, these results show that autoregressive action-token spaces are not merely compressed trajectory representations, but controllable semantic interfaces for human-guided robot behavior during inference.

\section{Related Work}
\label{sec:relatedwork}

\subsection{Diffusion Policy Steering}

Recent work has explored modifying pretrained robot policies at inference time without retraining the underlying model~\citep{du_dynaguide_2025, wagenmaker_steering_2025, wang_inference-time_2024}. 
Within diffusion-based robot policies~\citep{chi_diffusion_2023}, several methods steer generated trajectories by intervening on the denoising process itself. 
DSRL introduces reinforcement learning in latent noise space to guide frozen diffusion policies toward desired behaviors without updating the underlying policy parameters~\citep{wagenmaker_steering_2025}. 
Similarly, DiSCo formulates steering for diffusion policies by conditioning generation on partial human demonstrations and inpainting the remainder of the trajectory sequence~\citep{wang_disco_2026}. 
Related work such as DynaGuide further studies online guidance mechanisms that adapt policy behavior at inference time while preserving pretrained capabilities~\citep{du_dynaguide_2025}. 

In contrast, TS targets autoregressive VLAs that generate discrete action tokens sequentially. 
Rather than modifying latent trajectories during denoising, TS directly injects user-derived tokens into the autoregressive action stream and lets the policy complete the remaining sequence conditioned on the modified prefix. 
TS requires no retraining, auxiliary objective, or learned correction module.

\subsection{Shared Autonomy and Human Guidance}

Shared autonomy aims to combine human intent with autonomous robot capabilities in order to improve robustness, usability, and task success in uncertain environments~\citep{javdani_shared_2015, reddy_shared_2018, dragan_policy-blending_2013, jeon_shared_2020}. 
Classical shared autonomy systems often blend human and autonomous actions through arbitration mechanisms or assistance policies that infer user intent online~\citep{javdani_shared_2015, dragan_policy-blending_2013}. 
More recent approaches leverage deep learning to learn assistive behaviors directly from interaction data, enabling robots to adaptively support users during manipulation and navigation tasks~\citep{reddy_shared_2018}.

Language-guided correction has emerged as another promising direction for human-in-the-loop robot control~\citep{karamcheti_lila_2022, cui_no_2023}. 
Human-interaction steering methods enable online corrective feedback by mapping user inputs into policy adjustments during manipulation~\citep{wang_inference-time_2024}. 
Similarly, interactive robot learning and shared control methods have explored incorporating online human corrections through physical interventions, residual adjustments, or iterative preference feedback to refine policy behavior dynamically~\citep{jeon_shared_2020, reddy_shared_2018, kelly_hg-dagger_2019}. 
These approaches typically require either learned correction modules, additional finetuning, or specialized interfaces for translating human guidance into policy updates.

TS differs from prior interactive control systems in both representation and mechanism. Rather than learning a separate correction policy or latent controller, TS directly edits the pretrained VLA's native action-token stream during autoregressive generation. User inputs are converted into the same tokenized action representation used internally by the policy, allowing lightweight inference-time steering without retraining or overriding the underlying VLA.

\section{Methods}
\label{sec:methods}

\subsection{FAST action tokenization in autoregressive VLAs}

VLAs are built on top of large pretrained VLMs~\citep{driess_palm-e_2023, brohan_rt-2_2023, kim_openvla_2024} that have latent understanding across vision and language. 
VLMs are finetuned with robot trajectories alongside their visual and language observations such that understanding between tasks and environments become learned.
In autoregressive VLAs, robot actions are tokenized and these action tokens are appended to the existing VLM's token vocabulary.
This is distinct from a diffusion and flow VLA~\citep{black_pi_0_2024, intelligence_pi_05_2025, wen_tinyvla_2025} which generates action chunks without action tokens, and instead directly in action-space, initialized from noise.

Autoregressive VLAs therefore require a discrete representation of continuous robot trajectories in order to model actions using next-token prediction~\citep{janner_trajectory_2021, shafiullah_behavior_2022}. Traditional action tokenization methods discretize trajectories by quantizing actions independently at each timestep~\citep{brohan_rt-1_2023, brohan_rt-2_2023, kim_openvla_2024}. However, adjacent robot actions are often highly correlated and vary smoothly over time, making purely temporal discretization inefficient for autoregressive modeling.

$\pi_0$-FAST instead represents robot trajectories in the frequency domain~\citep{pertsch_fast_2025}. Given an action chunk
$
\mathbf{A} \in \mathbb{R}^{H \times D},
$
where $H$ denotes the action horizon and $D$ denotes the action dimension, FAST first normalizes the trajectory and applies a discrete cosine transform (DCT)~\citep{pertsch_fast_2025} independently across the temporal dimension of each action coordinate:
\begin{eqnarray*}
\hat{\mathbf{A}}_{k,d}
=
\sum_{t=0}^{H-1}
\mathbf{A}_{t,d}
\cos
\left(
\frac{\pi}{H}
\left(t+\frac12\right)
k
\right).
\end{eqnarray*}

The transformed coefficients, $\hat{\mathbf{A}}_{k,d}$ are quantized before flattening and ordering them from low to high frequency. The resulting coefficient sequence is then compressed into discrete autoregressive tokens using byte-pair encoding (BPE)~\citep{gage_new_1994}.

\subsection{Token Steering Formulation}

Token Steering (TS) steers autoregressive VLA policies by replacing a subset of generated FAST action tokens with tokenized user-induced perturbations, after which the policy completes the remaining sequence autoregressively (see Figure~1a). 
Formally, let $\bm{o}_t$ denote the visual observation at time $t$, $\ell$ denote the language instruction, $\bm{q}_t$ denote the robot's joint positions, and $\bm{A}_t \in \mathbb{R}^{H \times D}$ denote an action chunk at time $t$. 

An autoregressive VLA models action-token chunks $z_{1:N}$ with a learned policy $p_\theta$: 
\begin{eqnarray*}
p_{\theta}(z_{1:N} \mid \bm{o}_t, \ell, \bm{q}_t) = \prod_{i=1}^{N} p_{\theta}(z_{i} \mid z_{<i}, \bm{o}_t, \ell, \bm{q}_t).
\end{eqnarray*}
TS first records a low-dimensional user's intent $\bm{u} \in \mathbb{R}^6$.
In this study, $\bm{u}$ encodes one of six keyboard directions: left, right, back, forward, up or down.
This input is translated into a Cartesian velocity by scaling the directional vector $\bm{u}$ with a magnitude $m$, so that $\bm{v}_\textrm{cart} = m\bf{u}$.
The Cartesian velocity is converted to robot joint velocity through inverse kinematics: $\bm{\dot{q}}=\bm{J}(\bm{q})^\dagger \bm{v}_{\textrm{cart}}$, where $\bm{J}(\bm{q})^\dagger$ denotes the Moore-Penrose pseudoinverse of the manipulator Jacobian matrix.
Finally, the joint velocity $\bm{\dot{q}}$ is padded to the action horizon $H$ to create $\bm{\tilde{a}_{1:H}}$ which is then FAST-tokenized: $\tilde{z}_{1:K}=E_{\text{FAST}}(\bm{\tilde{a}_{1:H}}).$

TS inserts only a subset of steering tokens into the autoregressive buffer, defined by start index $b$ and injection window size $w$.
This limited intervention preserves the pretrained policy's ability to autoregressively synthesize temporally coherent and dynamically feasible trajectory continuations conditioned on the modified prefix:
\begin{eqnarray*}
z_i \sim \begin{cases} 
      \delta_{\tilde{z}_{i}}(\cdot) & i \in [b, b+w) \\
      p_{\theta}(\cdot \mid \tilde{z}_{<i}, z_{<i}, \bm{o}_t, \ell, \bm{q}_t) & i \notin [b, b+w) 
   \end{cases}.
\end{eqnarray*}

Within the injection window, TS deterministically fixes each action token to the injected steering token $\tilde{z}_i$, represented probabilistically as a Dirac point mass $\delta_{\tilde{z}_{i}}(\cdot)$. 
Outside of this window, tokens are autoregressively sampled from $p_\theta$.

\subsection{Robot embodiment and setup}

We use the DROID robot platform~\citep{khazatsky_droid_2025}. 
This comprises a Franka Panda arm fitted with the Robotiq 2F-85 gripper with $8$ degrees of freedom. 
We use one ZED scene camera, placed behind the manipulator to the left, and one ZED mini wrist camera, placed beneath the gripper. 
All experiments and tasks were evaluated with Physical Intelligence's $\pi_0$-FAST model trained on the DROID dataset with an action horizon of 10 and effective action dimension of 8. 
We used this model zero shot with no additional training or fine-tuning in all of our tasks.
Model inference was performed on a single GPU (NVIDIA GeForce RTX 5090).

\section{Experiments}

Our experiments evaluate: (1) how many tokens TS should inject, (2) which tokens (low vs high frequency) most effectively steer the policy, (3) if TS degrades policy diversity, (4) if TS can rectify incorrect language grounding, and (5) whether users can use TS to improve real-world task performance over an autonomous VLA baseline. Refer to Appendix~\ref{appendix:experimenttasks} for experiment setups.

\subsection{How many tokens do we inject? A tradeoff between small and large injection windows.}

\input{table1}

We investigated how the number of injected tokens affected downstream policy roll-out. 
We used FAST-tokenized action chunks $\bm{A} \in \mathbb{R}^{H \times D}$ with $H=10$ and $D=8$.
To evaluate this, we used a simple pick-and-place task. 
We positioned a bowl beneath the manipulator, with four colored blocks arranged around the bowl along the left, right, forward, and backward directions relative to the robot base.
Each trial was given an intentionally ambiguous language instruction that was directionally agnostic: pick up the block and place it in the bowl.
To steer the policy towards a particular block, we injected tokens for 1-second velocity at a velocity of $1$ m/s towards one of the four directions (up, down, left, right). 
Empirically, we observed full sequence lengths of 8 to 36 tokens (including two EOS tokens).
Thus, we experimented with seeding from 1 to 6 tokens.
The injection window for the generated action token sequence consisted of the first $k$ steering tokens.
We swept across an injection window size $\in \{1, 2, 4, 6\}$.
We define the steering intent rate (SIR) as the proportion of trials in which the arm picks up and places the block in the intended direction (see Appendix~\ref{appendix:metricssir}).
We define the mean path efficiency (MPE) as the straight line path length from the end-effector to the block divided by actual path length, averaged across successful trials (see Appendix~\ref{appendix:pathefficiency}).

For trials where only a single steering token, $\tilde{z}_0$, was injected, the robot's trajectory often went to an incorrect block (Figure~\ref{fig:expt1}a, red trajectories).
The first token did not contain enough broad directional information to steer the policy to align with the injected intent. 
However, as we injected more tokens into the executed action buffer, we observed a higher alignment with the injected intent (Figure~\ref{fig:expt1}a, blue trajectories). 
We also observed a decrease in path efficiency as the injection window increased (Figure~\ref{fig:expt1}b), suggesting that larger injections reduce the policy's role in refining the trajectory.
In practice, we noticed overshooting of the block for higher injection window sizes.
Thus, there is an important tradeoff for token injection: overall, small injection windows reflect greater policy influence while larger injection windows reflect greater steering input influence, but potentially with inefficient and suboptimal actions. 

\input{expt1-fig}

\subsection{Low-frequency FAST tokens more effectively steer the policy}

\input{expt2-fig}

We next investigated whether editing low-frequency tokens is preferred to editing high-frequency tokens. 
In the FAST tokenizer algorithm, tokens are ordered from low to high frequency representations before BPE.
This ordering induces a hierarchical semantic structure over the autoregressive token sequence.
Early FAST tokens correspond to low-frequency trajectory components that strongly influence global motion semantics, while later tokens encode progressively finer local trajectory details.
As a result, autoregressive generation over FAST tokens can be interpreted as progressively refining a trajectory from coarse global motion structure to high-frequency corrections.

To assess which tokens led to empirically steerable policies, we fixed an injection window of size $2$ and injected tokens using a start index $b$ swept across $b \in [0, 3]$. 
We then performed the same experiment described in Section 4.1. 
We observed that injecting low frequency tokens resulted in significantly higher steering success compared to high frequency tokens (Figure~\ref{fig:expt2}a,b). 
As we shifted the injection window towards higher frequency tokens, we observed a degradation in steering intent: the steering input mattered less when injected in high-frequency tokens. 
Notably, task progression did not degrade: the policy still picked and placed a block, but the steering intent success rate (picking the block aligned with the steering input) degraded. 
With an injection window of just two tokens, the policy is still able to recover from injected tokens regardless of where they are placed and align itself to the language instruction of picking up the block and placing it in the bowl.

\subsection{Token Steering is out-of-distribution: does it lead to a collapse in action entropy?}
\input{expt3-fig}

TS intentionally injects tokens that the policy did not generate itself, creating an out-of-distribution intervention in the autoregressive action stream.
While steerability and control of a robot policy is desired, there is a possibility that injecting action tokens may negatively impact VLA performance by limiting the policy's ability to take diverse actions based on its environment.
For example, to grasp a randomly oriented block, the policy should be able to rotate the end effector to align with the block's orientation, increasing the chances of a successful pick by picking it on its edges rather than its corners.
TS should not negatively impact the generation of diverse future VLA actions: the policy should be able to autoregressively execute actions to successfully pick up blocks of any orientation following token injection.

To assess this, we evaluated whether token injection causes entropy collapse of the action distributions following token injection.
We performed an experiment where we generated 20 different visual scenes of randomly scattered blocks.
We then injected two steering tokens in each trial, and then autoregressively generated the remaining trajectory action tokens for each steering direction (left, right, back, forward).
We recorded both the total token diversity across each direction as well as the distribution of logits for each autoregressive step post-seeding.
We found that while the entropy (refer to Appendix~\ref{appendix:entropy}) of the 3rd token, immediately following the injected tokens, is relatively small (Figure~\ref{fig:expt3}b), this is a temporary effect: following the 3rd token, the next token entropy significantly increases until later on in the trial, when the policy is confident in the next action token. 
Injecting steering tokens may therefore temporarily shift the expert out of distribution, but the policy quickly recovers and continues generating diverse downstream trajectories.
These results suggest that TS biases trajectory generation without collapsing the policy's autonomous adaptability.

\subsection{Token Steering is out-of-distribution: can it overcome incorrect language following?}

\input{expt4-fig}

We next investigated whether the same out-of-distribution intervention introduced by TS can help correct policy failures during language grounding.
Instead of treating out-of-distribution steering purely as a robustness concern, we ask whether it can beneficially redirect trajectories when the autonomous policy selects the wrong object.

We performed an experiment where we placed four blocks around a bowl: a blue cube, a green cube, a longer blue rectangular prism, and a longer green rectangular prism.
For each trial, the language prompt specified which color and type of block the robot should pick up and place in the bowl.
We observed that $\pi_0$-FAST had imperfect language following (Figure~\ref{fig:expt4}b): its success rate in following the language commands ``pick up the green cube and place it in the bowl'' and ``pick up the blue rectangular block and place it in the bowl'' was $0\%$ as the policy always picked up the incorrect object.

We then injected steering tokens toward the correct object during roll-out.
TS redirected the policy towards the correct object and achieved $100\%$ task success on both failure cases, as shown qualitatively in Figure~\ref{fig:expt4}a and quantitatively in Figure~\ref{fig:expt4}b,.
The percentage of steering token usage was relatively small (Figure~\ref{fig:expt4}c): less than $25\%$ of the total trajectory's tokens had to be steered to overcome incorrect language following.
These results suggest that out-of-distribution token injection is not merely tolerated by the autoregressive policy.
Instead, it provides a controllable mechanism for correcting policy behavior while preserving the pretrained policy's downstream trajectory generation capabilities.

\subsection{Users can steer VLAs to achieve better performance than autonomous policies}

To evaluate the potential for real-world task applicability, we performed a user study with seven users on two household tasks to assess the interplay between steering and generalization.
Users injected FAST tokens via a keyboard, injecting a magnitude of 0.5 m/s that was then transformed into a sequence of steering tokens (injection window = 4) corresponding to the given direction.
Velocity and injection window parameters were chosen based on task environment for ideal control of the system.
User intent was represented using six discrete Cartesian motion primitives corresponding to forward, backward, left, right, up, and down keyboard commands.
All experiments were approved by the institutional IRB.

\input{table2}

\textbf{Task 1: put the banana in the drawer (Appendix~\ref{appendix:userstudies}).} 
The first task evaluated whether the vanilla $\pi_0$-FAST VLA could benefit from steering tokens in order to successfully complete a task requiring precision and environmental composition.
Users were tasked with picking up a toy banana and placing it in an open drawer; they were then instructed to assist the robot with closing the drawer by sliding it shut.
Trials ran either until the task was successful, the time-out of 800 timesteps was reached, or the trajectory entered an unrecoverable failure mode. 
As shown in Table~\ref{tab:userstudy}, the baseline performance of $\pi_0$-FAST achieved a $10.0\%$ task success rate.
Common failure modes included pushing the drawer from a suboptimal angle which led to jamming the drawer, moving the drawer backwards until it fell out of the slot, and being unable to locate the drawer itself.
When users steered the policy with keyboard inputs, TS increased the success rate from $10.0\%$ to $72.5\%$ and reduced median completion time from $74.0$~s (autonomous) to $42.0$~s (token steering) (Table~\ref{tab:userstudy}). Wilcoxon rank-sum tests showed significant improvements in both Banana Drawer success rate ($p=0.003$) and completion time ($p=0.020$).
Notably, we empirically found that users were able to quickly navigate the gripper behind the drawer after it dropped the banana in place, and situate the gripper into an optimal position. 
Users then prompted the gripper forward to close the drawer gently.

A key observation from the drawer-closing segment was that users and the policy jointly produced safe behavior.
Users provided only low-dimensional keyboard inputs corresponding to fixed Cartesian velocities.
A purely manual controller using these inputs would likely close the drawer harshly.
In contrast, TS biased the trajectory direction while the policy autoregressively generated later tokens that refined higher-order motion details.
Because the policy still attended to visual and language context, it moved forward more slowly and precisely, allowing the robot to close the drawer gently.

\textbf{Task 2: Swap the sponges (Appendix~\ref{appendix:userstudies}).}
The next task we performed injects user intent in a scenario where autoregressive VLAs often fail: tasks requiring memory. 
Critically, $\pi_0$-FAST does not have memory of past observations or actions.
In this context, TS allows the user to manage the high-level completion of the task without explicitly handling all of the low-level dexterity.

For this task, we placed three plates from left to right collinearly in front of the robot base.
Two sponges---one blue and one orange---were placed on two of the plates.
Users were asked to swap the two sponges by using the third empty plate as a temporary spot.
Using steering tokens, users guided the arm to the desired sponge until the task was complete.
Trials ran for a maximum of four minutes.

The autonomous policy completed this task successfully $0\%$ of the time.
However, it achieved a task progression rate of 16.7\% (Table~\ref{tab:userstudy}).
Appendix~\ref{appendix:userstudies} describes how we evaluated the swap sponges task progression success rate.
Common failure modes---aside from the inherent memory issue---included dropping sponges outside of the plates and being unable to recover them, moving the plates out of line, and staying stagnant in front of the plates.
Users were able to recover from these errors robustly, achieving an average success rate of $93.8\%$ (Table~\ref{tab:userstudy}), which was statistically significant under a Wilcoxon rank-sum test ($p<0.001$).
If sponges were dropped, users were able to retrieve them and move them to the desired position.
Additionally, users were able to keep track of sponge start state and accurately swap them within four minutes.
This demonstration of steerability is an important novelty that both displays support for user intent as well as general task completion from a task the VLA failed to complete in our trials. 

\section{Discussion}
\label{sec:conclusion}

Autoregressive VLAs provide a unified architecture combining visual, language, proprioceptive, and action representations into a single token-based architecture.
We find that injecting a steering signal in this native token space provides an intuitive interface for users to exert directional influence in an autonomous trajectory.
Because TS operates zero-shot at inference time, it can apply across tasks without collecting new data or finetuning the policy.

Although we use keyboard inputs, TS does not depend on keyboards.
Any low-dimensional intent signal that can map to a velocity command could generate steering tokens.
This makes TS especially relevant for brain-computer interfaces (BCIs), where current systems have demonstrated impressive reach-and-grasp and prosthetic-arm control but still face challenges in providing high-dimensional, dexterous, continuous control~\citep{hochberg_reach_2012, collinger_high-performance_2013}. Recent BCI work also shows the strength of decoding structured low-dimensional intent suggesting that low-dimensional systems can be powerful when paired with an appropriate generative interface~\citep{willett_high-performance_2021}.

\subsection{Limitations}

Our autoregressive steering, while real-time, is naturally slower than a flow- or diffusion-based VLA since it is constrained by the nature of next-token prediction latency.
In tasks requiring quick manipulation or time-sensitive handling, a flow or diffusion based VLA may be more ideal.
Work has been done to amortize inference time by supporting asynchronous action chunk generation~\citep{black_rtc_2025}, but the scope of this work did not explore combining steerability with asynchronous action chunking. 
Additionally, improving inference speed is still an open problem for autoregressive VLAs, and presumably will improve conditionally with hardware improvements.

\section*{Author contributions} 
J.C. conceived and implemented Token Steering, conducted the experiments and user study, performed the analyses, generated the figures, and wrote the paper.
J.C.K. supported experiment ideation and participated in paper review and editing. 

\section*{Acknowledgments}
This work was supported by NIH DP1HD121548, DP2NS122037, and the UCLA-Amazon Science Hub award (all to J.C.K).

\section*{Competing Interests}
J.C.K. is the inventor of intellectual property owned by Stanford University that has been licensed to Blackrock Neurotech and Neuralink Corp.
J.C.K. has a provisional patent application related to AI copilots for brain--computer interfaces that is owned by the Regents of the University of California. 
J.C.K. is a co-founder of Luke, is on its Board of Directors and has a financial interest in it. 
The other authors declare no competing interests.

\newpage

\clearpage

\bibliography{references}

\clearpage
\appendix

\section*{Appendix}

\section{Experiment Tasks and Setup}
\label{appendix:experimenttasks}

For our characterization experiments and user studies, we defined five different environmental setups.

\input{appendixA}

Figure~\ref{fig:appendixA} shows the five workspace configurations used across our characterization experiments and user study tasks. Panels (a)--(c) were used for controlled token-steering analyses, while panels (d)--(e) were used for real-world user studies. All setups used the same DROID robot embodiment, camera configuration, and $\pi_0$-FAST policy described in Section~\ref{sec:methods}.

\subsection{Characterization Experiments}

\textbf{(1) Injection window experiment.} A dark blue bowl was placed directly under the gripper. Four colored blocks were placed around the central bowl along the left, right, forward, and backward directions relative to the bowl. The policy received an intentionally direction-agnostic instruction, ``pick up the block and place it in the bowl.'' 
Steering tokens were injected toward one target block for 1-second. The tokenized Cartesian velocity was 1.0 m/s.

\textbf{(2) Frequency tokens experiment.} Same block-and-bowl layout as previous experiment, but used to evaluate how the injection start index affected steering success when editing lower-versus higher-frequency FAST tokens. Tokens were injected for 1-second at 1.0 m/s.

\textbf{(3) Out-of-distribution tokens experiment.} Multiple colored blocks were randomly scattered around the workspace to evaluate whether token injection preserved diversity in the autoregressively generated downstream tokens. A total of 20 random scenes were used across 80 trials. Tokens were injected for 1-second at 1.0 m/s. The injection window size was 2 with a start index of 0. The policy was instructed to ``pick up the block`` via its language instruction.

\textbf{(4) Overcoming policy actions during incorrect language following.} Similar to experiments 1 and 2, a bowl was placed directly under the gripper. Four blocks were placed around the bowl: a blue rectangular block, a green rectangular block, a blue cube, and a green cube. A tokenized 0.5 m/s Cartesian velocity was injected in the case of incorrect language following. The injection window size was 2 with a start index of 0. We swept across 40 trials, with the language instruction cycling through  ``pick up the blue cube and place it in the bowl``, ``pick up the green cube and place it in the bowl``, ``pick up the blue rectangular block and place it in the bowl``, and ``pick up the green rectangular block and place it in the bowl.``

\subsection{User Studies}
\label{appendix:userstudies}

Users were given up to 3 practice trials to familiarize themselves with the system before starting each task. We found empirically that the task required not a fixed injection, but instead a sequence of short bursts. Given the different behavior of the task, it necessitated a different set of hyperparameters. Thus, we used a tokenized Cartesian velocity of 0.5 m/s for each key-press, with a token injection window of 4 tokens.

\textbf{(1) Drawer Task} A toy banana and open drawer were placed in the workspace. Users were instructed to pick up the banana and place it in the drawer, then close the drawer. 

\textbf{(2) Sponge Swapping} Three plates were arranged in a row with the edge of the plate 22 cm from the base of the robot. Two colored sponges were placed on separate plates, with each trial seeing a different permutation. Users guided the policy to swap the sponge locations using the empty plate as temporary storage spot. The task progress was measured in 3 stages, each stage was considered successful if the user picked up and placed a sponge in the correct spot for a total of three movements.

\section{Performance Metrics}
\label{appendix:metrics}

We evaluated our system using several metrics designed to quantify steering effectiveness, trajectory quality, policy diversity, and overall task performance.

\subsection{Steering Intent Rate}
\label{appendix:metricssir}

For experiments involving directional steering, we measured \textit{steering intent rate} (SIR), defined as the percentage of trials in which the robot selected the object corresponding to the injected steering direction. 
For example, in the block-selection experiments, a trial was considered successful if the robot picked up the block located in the same directional quadrant as the injected steering tokens.

\subsection{Path Efficiency}
\label{appendix:pathefficiency}

To quantify trajectory smoothness and efficiency, we computed the \textit{path efficiency} of the robot end-effector trajectory:
\begin{eqnarray*}
\text{Path Efficiency}
=
\frac{d_{\text{straight}}}{d_{\text{traj}}}
\end{eqnarray*}
where $d_{\text{straight}}$ denotes the Euclidean distance between the trajectory start and end positions, and $d_{\text{traj}}$ denotes the cumulative distance traveled by the end effector throughout the rollout.
Higher values correspond to straighter and more efficient trajectories, while lower values indicate less efficient motion with greater curvature or overshooting.

For experiment 1, we report the mean path efficiency (MPE) averaged across all successful trials.

\subsection{Task Success Rate}
\label{appendix:tasksuccessrate}

For user-study experiments, we measured \textit{task success rate}, defined as the percentage of trials in which the full task objective was completed successfully before timeout or unrecoverable failure.

For the drawer-closing task, a trial was considered successful if:
\begin{enumerate}
    \item the banana was placed into the drawer, and
    \item the drawer was fully closed.
\end{enumerate}
For the sponge-swapping task, a trial was considered successful if the two sponges exchanged positions using the third plate as temporary storage within the allotted time limit. This requires three total sponge movements. Each individual sponge movement (pick up and place) attributed to $\frac{1}{3}$ of the success rate for a total of 100\% for trials with fully swapped sponges.

\subsection{Task Completion Time}
\label{appendix:taskcompletiontime}

We recorded \textit{task completion time} as the elapsed wall-clock time from rollout initialization until successful task completion.
For unsuccessful trials, completion time was not included in the reported statistics.
We report the median completion time across successful trials due to the presence of outlier rollouts with prolonged recovery behavior.

\subsection{Autoregressive Token Entropy}
\label{appendix:entropy}

To evaluate whether token injection constrained downstream policy diversity, we measured the entropy of the autoregressive action-token distributions following steering-token insertion.

For an autoregressive token distribution $p(z_i)$, entropy was computed as:
\begin{eqnarray*}
H(z_i)
=
-\sum_{z_i} p(z_i)\log p(z_i).
\end{eqnarray*}

Entropy was measured for each generated token after the injected token window.
Higher entropy indicates greater diversity in the policy's generated token distribution, while lower entropy reflects higher confidence in a particular continuation.

\section{Token Diversity After Seeding}
\label{appendix:tokendiversity}

\input{appendixdiversity}

\end{document}

%% file: fig1.tex
\begin{figure}[t!]
    \centering
    \includegraphics[width=1.00\linewidth]{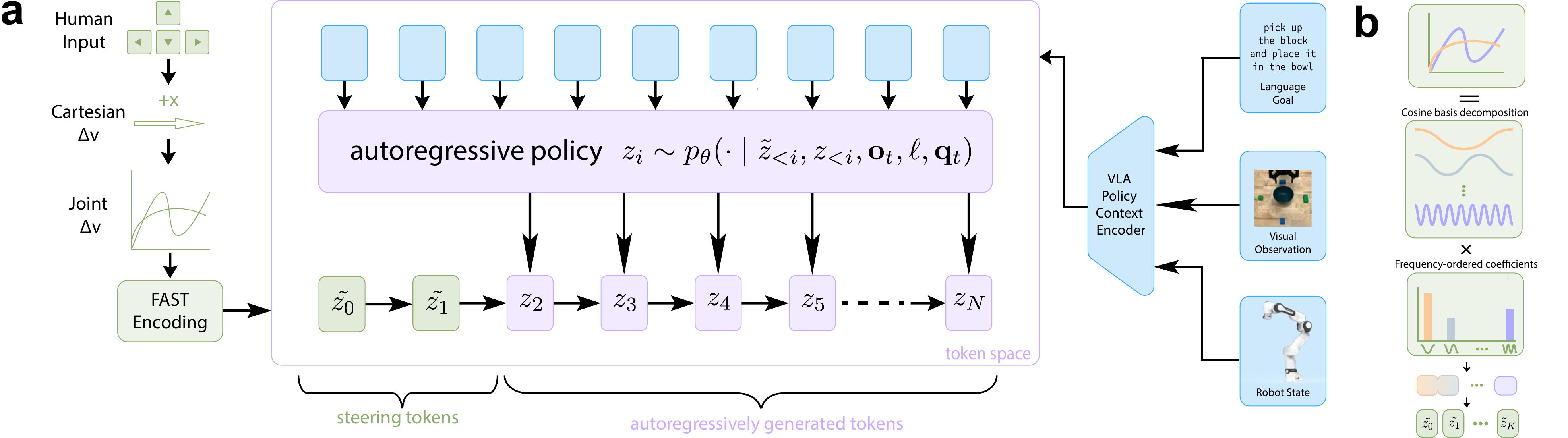}
    \caption{
Overview of Token Steering (TS).
(a) A low-dimensional user input is converted into Cartesian velocity, transformed into joint velocity, FAST-tokenized, and injected into the autoregressive action-token sequence of a pretrained VLA policy. The policy then autoregressively completes the remaining trajectory conditioned on the modified token prefix together with visual observations, language instructions, and robot state.
(b) FAST tokenization transforms joint-velocity-based action trajectories into frequency-ordered coefficients using a discrete cosine transform (DCT), producing low-frequency tokens that encode coarse trajectory structure and high-frequency tokens that encode fine motion refinements.
}
    \label{fig:fig1}
\end{figure}

%% file: table1.tex



\begin{wraptable}{r}{0.42\textwidth}

\vspace{-1.2em}

\centering

\small

\begin{tabular}{c|cc}

\toprule

Injection Window & SIR $\uparrow$ & MPE $\uparrow$ \\

\midrule

1 & 0.38 & 0.724 \\

2 & 1.00 & 0.663 \\

4 & 1.00 & 0.602 \\

6 & 0.94 & 0.564 \\

\bottomrule

\end{tabular}

\vspace{-0.3em}

\caption{Effect of injection window size on steering performance.}

\label{tab:expt1}

\vspace{-1em}

\end{wraptable}

%% file: expt1-fig.tex
\begin{figure}
    \centering
    \includegraphics[width=1.0\linewidth]{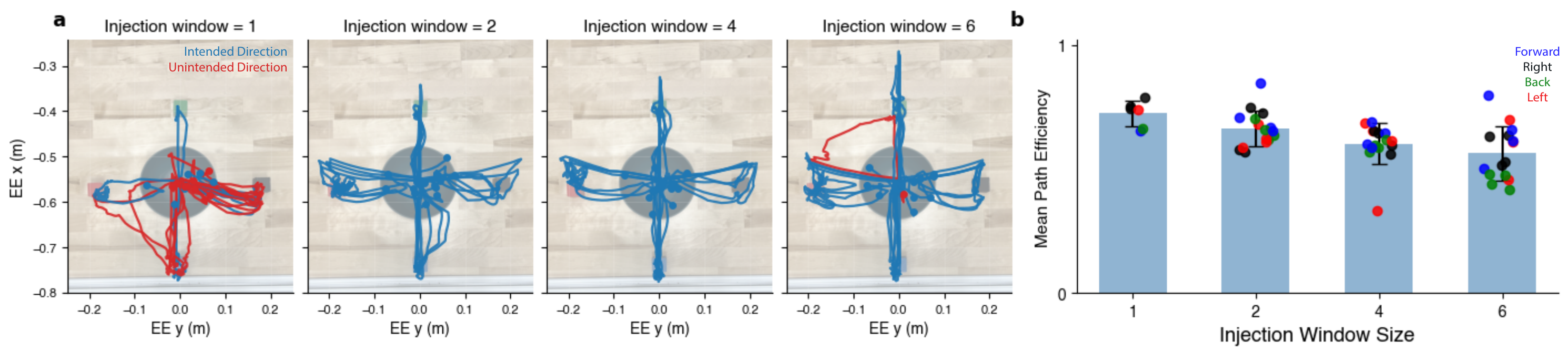}
    \caption{
Effect of injection window size on steering behavior.
(a) End-effector trajectories for different steering-token injection window sizes. 
(b) Mean path efficiency across injection window sizes.
}
    \label{fig:expt1}
\end{figure}

%% file: expt2-fig.tex
\begin{figure}[b!]
    \centering
    \includegraphics[width=1.0\linewidth]{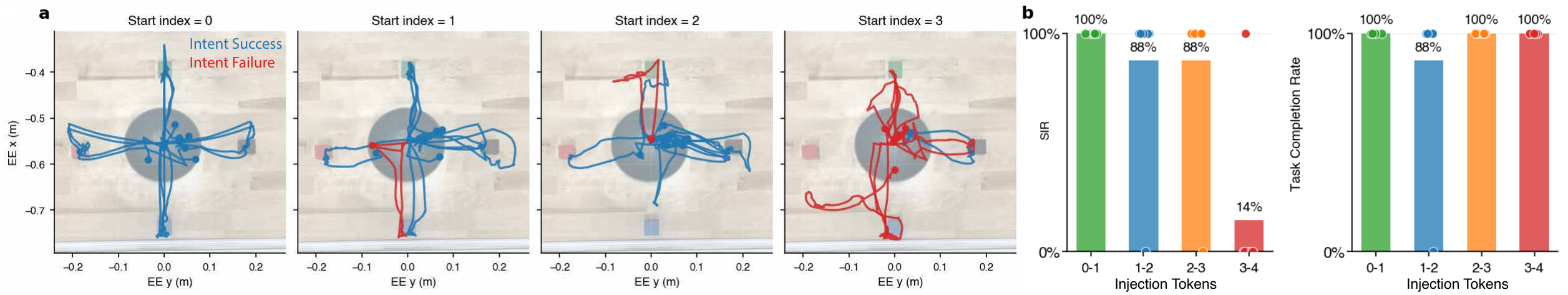}
    \caption{
    Influence of FAST token frequency on steering effectiveness.
    (a) Example trajectories produced when injecting steering tokens at different positions in the FAST token sequence.
    (b) Steering intent success rate and overall task completion rate as a function of injection start index.
    }
    \label{fig:expt2}
\end{figure}

%% file: expt3-fig.tex
\begin{figure}
    \centering
    \includegraphics[width=1.0\linewidth]{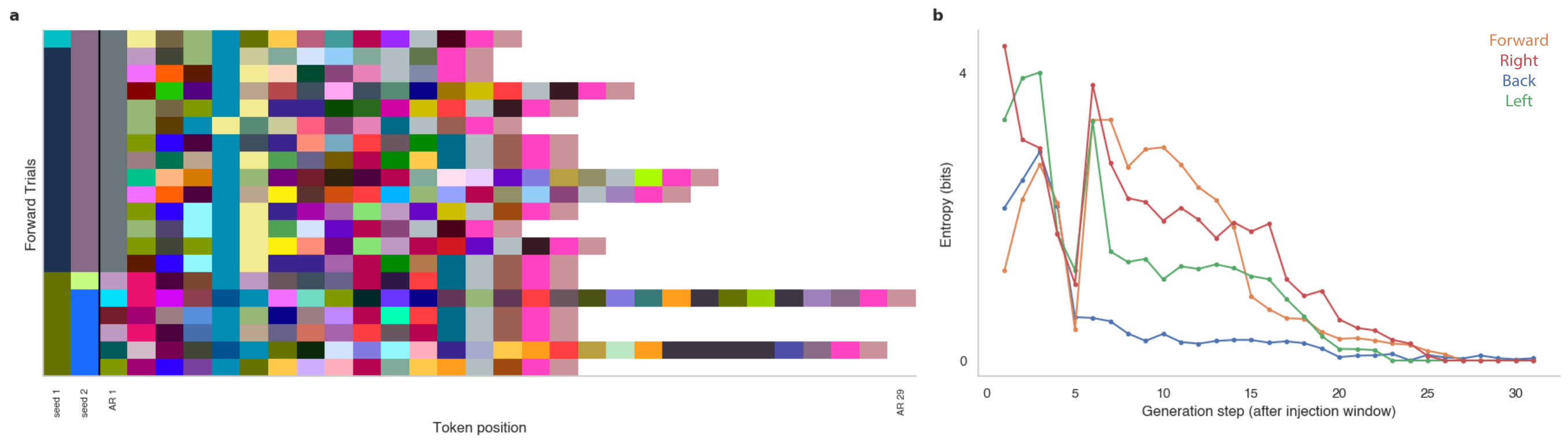}
    \caption{
    Out-of-distribution behavior following steering-token injection.
    (a) Generated autoregressive token sequences after injecting identical steering-token prefixes across multiple scenes and directions.
    (b) Entropy of the autoregressive token distribution after steering-token insertion.
    }
    \label{fig:expt3}
\end{figure}

%% file: expt4-fig.tex
\begin{figure}[b!]
    \centering
    \includegraphics[width=1.0\linewidth]{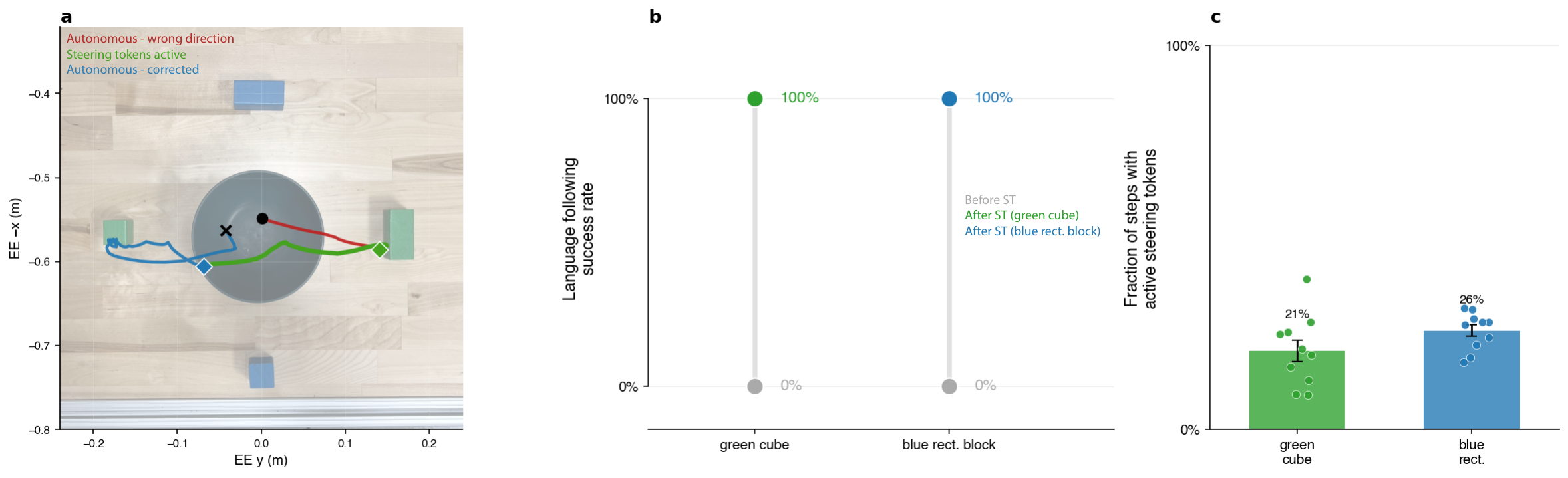}
    \caption{(a) Example correction for a trajectory where the intended block was the green cube. (b) Success rates of language following before and after TS. (c) Steering tokens were used for a minority of the entire trajectory for both failure cases.}
    \label{fig:expt4}
    \vspace{-1em}
\end{figure}

%% file: table2.tex
\begin{wraptable}{r}{0.50\textwidth}
\vspace{-1.0em}
\centering
\small

\begin{tabular}{l|cc}
\toprule
Task & Success & Time (s) \\
\midrule
Drawer w/o Steering & 10.0 & 74.0 \\
Drawer w/ Steering & 72.5 & 42.0 \\
Swap w/o Steering & 16.7 & -- \\
Swap w/ Steering & 93.8 & 133.8 \\
\bottomrule
\end{tabular}

\vspace{-0.2em}
\caption{
Success rates (\%) and median completion times for both tasks under autonomous and steering-assisted conditions.
}
\label{tab:userstudy}

\vspace{-1.2em}
\end{wraptable}

%% file: appendixA.tex
\begin{figure}[H]
    \centering
    \includegraphics[width=1.00\linewidth]{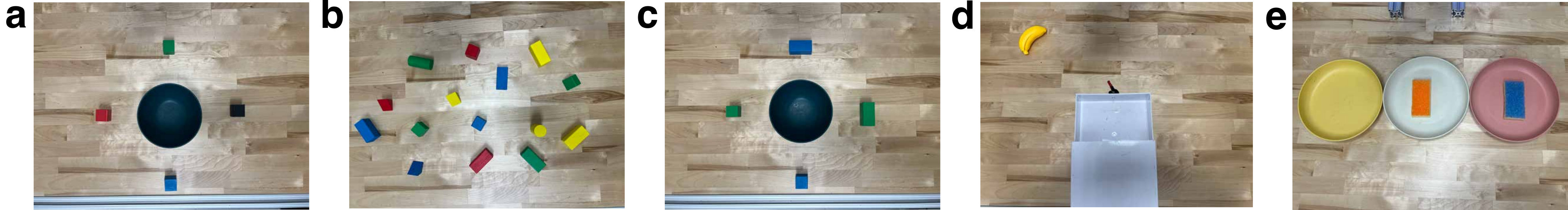}
    \caption{
    Workspace configurations used across all experiments and user studies.
    (a) Injection-window experiment and token frequency injection experiment.
    (b) Out-of-distribution diversity experiment.
    (c) Language-grounding correction experiment.
    (d) Drawer-closing user study.
    (e) Sponge-swapping user study.
    }
    \label{fig:appendixA}
\end{figure}

%% file: appendixdiversity.tex
\begin{figure}[H]
    \centering
    \includegraphics[width=1.00\linewidth]{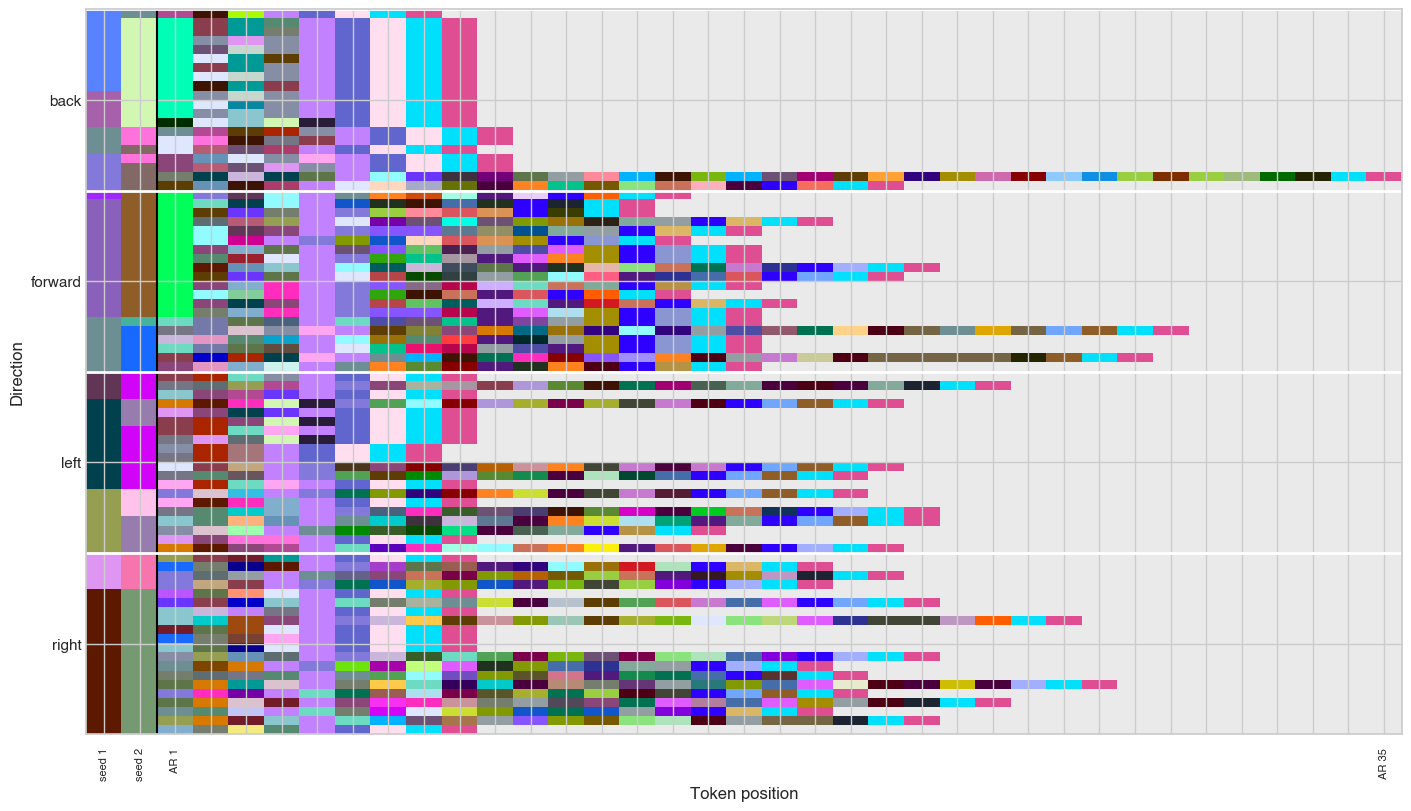}
    \caption{
    Autoregressive token diversity after injecting two steering tokens.
    Each row corresponds to a different steering direction across 20 randomized visual scenes. Although all trajectories share the same injected steering-token prefix, downstream autoregressive generations remain diverse, indicating that TS biases trajectory generation without collapsing policy variability.
    }
    \label{fig:appendixDiversity}
\end{figure}